\documentclass[letterpaper, 10 pt, conference]{ieeeconf}  

\IEEEoverridecommandlockouts                              

\usepackage{graphics} 
\usepackage{epsfig} 
\usepackage{subfig}
\usepackage{amsmath} 
\usepackage{amssymb}  
\usepackage{booktabs} 
\usepackage{cite}

\usepackage{algorithm}  
\usepackage{algorithmic} 

\title{\LARGE \bf
Learning to Segment and Represent Motion Primitives from Driving Data for Motion Planning Applications
}

\author{Boyang Wang$^{1,2}$,~\IEEEmembership{Student Member,~IEEE}, Jianwei Gong$^{1}$,~\IEEEmembership{Member,~IEEE}, Ruizeng Zhang$^{1}$, Huiyan Chen$^{1}$
\thanks{*This work was supported by the National Natural Science Foundation of China (No.91420203 and No.61703041, and the paper is also funded by International Graduate Exchanging Program of Beijing Institute of Technology}
\thanks{$^{1}$ All the authors are with the School of Mechanical Engineering, Beijing Institute of Technology, Beijing, China, 100081
        {\tt\small gongjianwei@bit.edu.cn}}%
\thanks{$^{2}$ Boyang Wang is also with the Interactive Digital Human group of CNRS-UM LIRMM, UMR5506, Montpellier, France, 34095.
        {\tt\small wbythink@hotmail.com}}%
}

\begin{document}

\maketitle
\thispagestyle{empty}
\pagestyle{empty}

\begin{abstract}

Developing an intelligent vehicle which can perform human-like actions requires the ability to learn basic driving skills from a large amount of naturalistic driving data. The algorithms will become efficient if we could decompose the complex driving tasks into motion primitives which represent the elementary compositions of driving skills. Therefore, the purpose of this paper is to segment unlabeled trajectory data into a library of motion primitives. By applying a probabilistic inference based on an iterative Expectation-Maximization algorithm, our method segments the collected trajectories while learning a set of motion primitives represented by the dynamic movement primitives. The proposed method utilizes the mutual dependencies between the segmentation and representation of motion primitives and the driving-specific based initial segmentation. By utilizing this mutual dependency and the initial condition, this paper presents how we can enhance the performance of both the segmentation and the motion primitive library establishment.  We also evaluate the applicability of the primitive representation method to imitation learning and motion planning algorithms. The model is trained and validated by using the driving data collected from the Beijing Institute of Technology intelligent vehicle platform. The results show that the proposed approach can find the proper segmentation and establish the motion primitive library simultaneously.

\end{abstract}

\section{Introduction}
In order to increase the potential for acceptance of intelligent vehicles, the human factors should be regarded as one of the crucial factors in system design\cite{martinez2018driving,schnelle2018feedforward,schnelle2017personalizable}. A commonly accepted concept to achieve such human-like behavior is imitation learning which enables the system to learn from the demonstrations and subsequently reproduce the learned skills to new situations\cite{argall2009survey,mulling2013learning,lioutikov2015probabilistic}. Regarding complex driving task as a sequence of motion primitives (MPs) provides two obvious benefits. First, the motion planning tasks will be simplified to desired points setting and the selection of MPs from the library. Second, the learned library of the MPs can also be reused to other driving tasks. These are essential for the motion planning system, as the system can be adapted to different situations by only replacing several point-to-point MPs within the trajectory sequence. A fundamental issue of such methods is acquiring the MPs autonomously without relying on the labeled driving data. In this paper, we propose a framework for segmenting unlabeled vehicle trajectory data into a library of MPs to solve this problem. Two issues need to be considered in the proposed framework, the segmentation of the observed driving trajectory and the representation of the segmented trajectory primitives.

\par The development of motion planning algorithms is crucial for the intelligent vehicle system. Most of the research groups apply interpolation or graph search to solve the motion planning problem in real implementations\cite{gonzalez2016review}. In the method of interpolation, the trajectories are often represented by clothoids\cite{vorobieva2014automatic}, Bezier curves\cite{petrov2014modeling} and polynomial curves\cite{rastelli2014dynamic}. Graph search is another algorithm which searches the optimal path considering several constraints\cite{urmson2008autonomous,howard2007optimal}. Urmson et al.\cite{urmson2008autonomous} proposed to define the motion as the longitudinal and lateral parameterized functions separately. The linear velocity profile takes the form of a constant profile, linear profile, linear ramp profile or a trapezoidal profile. The lateral part is defined as a second-order spline profile. While the above-mentioned representations are proved to be favorable and effective for optimized trajectory generation, there is no consensus that they are equally applicable to learning algorithms. By using the collected driving data, Schnelle et al.\cite{schnelle2017driver} proposed a combined driver model which includes a compensatory transfer function and an anticipatory component based on road geometry to generate the personalized driver's desired trajectory. However, this approach relies heavily on parameterized driving situations and therefore misses the generality to be chosen as the basic function to establish the motion primitive library. 

\par The representation approach which we used in this paper is based on the dynamical systems and called the Dynamic system movement primitives (DMPs). DMPs have many advantages over other representation methods, such as being robust against perturbations, allowing changing of the final position and time duration without altering the overall shape of the trajectory. Furthermore, DMPs are suitable for imitation learning\cite{ijspeert2002movement} and reward-driven self-improvement\cite{kober2009policy}, thus have been successfully applied to learn numerous motor skills in robotics\cite{ijspeert2013dynamical}. And the finitely varying joint angular values are utilized directly as the parameters of the DMPs in the field of robotics. However, in the driving situation, there are significant changes in the values of the velocity and course angle but limited changes in the relative variation. Considering the specificity of driving situation and differing from the traditional DMPs that widely used in the robotics, the parameters of the proposed MPs are defined by the initial conditions of each motion primitive and control increments within the time duration.  

\begin{figure*}[t]
\centering
\subfloat[A total of 81 hours of driving data are collected by the platform. The path data which are gathered from the integrated navigation unit are utilized in this paper.]{
\begin{minipage}[t]{0.28\textwidth}
\centering
\includegraphics[scale=0.45]{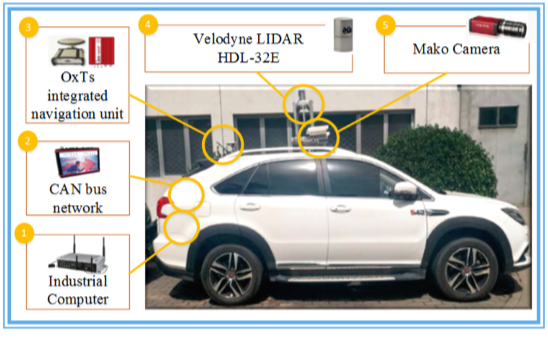}
\end{minipage}
\label{fig１(a)}
}
\hspace{10pt}
\subfloat[Sampled-data of one typical demonstration. The points show the initial segmentation points.]{
\begin{minipage}[t]{0.28\textwidth}
\centering
\includegraphics[scale=0.18]{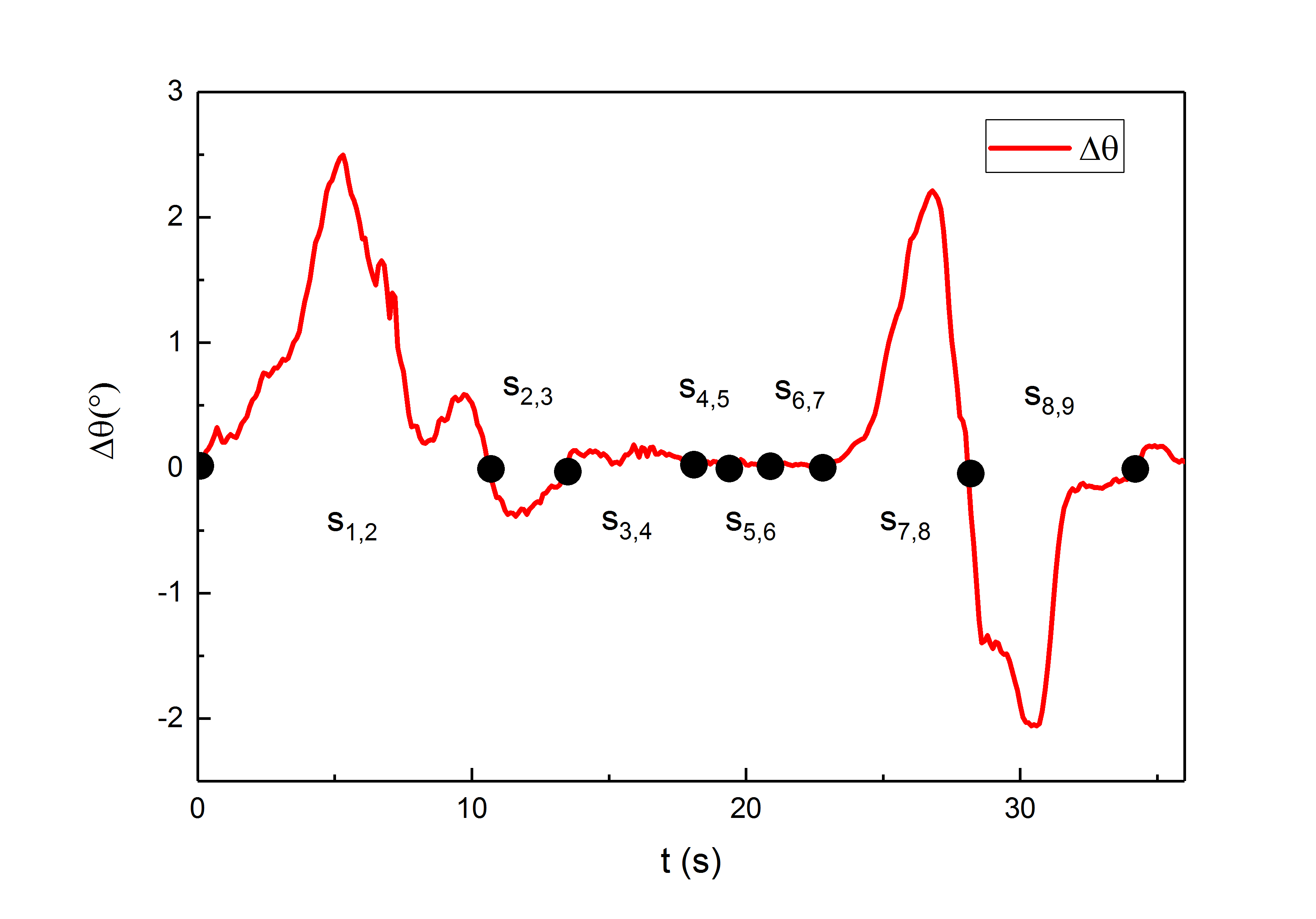}
\end{minipage}
\label{fig１(b)}
}
\hspace{10pt}
\subfloat[The result of final segmentation and learned MPs. The background color indicates the identified type of MPs.]{
\begin{minipage}[t]{0.28\textwidth}
\centering
\includegraphics[scale=0.18]{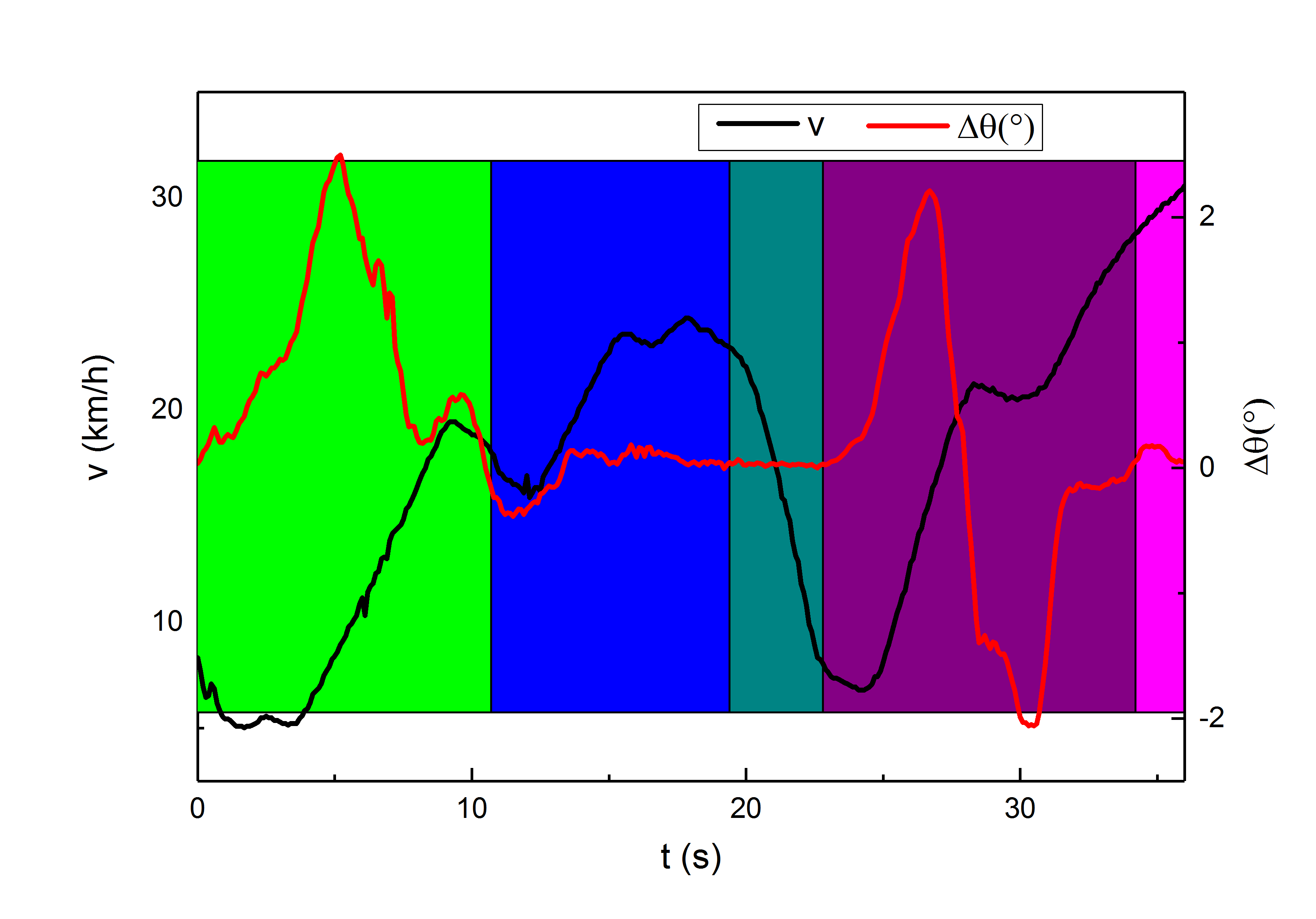}
\end{minipage}
\label{fig１(c)}
}
\caption{The overall flow of our framework. First, the trajectory data are collected by the Beijing Institute of Technology(BIT) intelligent vehicle platform (a). In order to maintain the integrity of the steering process, the initial cutting points are set based on the zero detection of the course deviation (b). According to the initial segmentation, the probabilistic segmentation method is applied based on the representation of the learned MPs. The false segmentation points are merged and the motion primitive library is established (c).}
\label{fig１}
\end{figure*}

\par Inferring driver behavior from the driving data is one of the crucial issues in the field of driving assistance system. Some researchers found the algorithms would become more efficient if the complex driving tasks could be decomposed into elemental movements. Li et al. established a driving behavior dataset of thirty-three participants by doing a field operational test around Beijing \cite{Li2015Field}. The dataset was then used to train the driving style estimation model by concentrating on the transition patterns between typical maneuver primitives. The typical maneuver episodes were segmented manually based on the specific rules designed by the author \cite{Li2017Estimation}. Bender et al.\cite{bender2015unsupervised} developed a two-level unsupervised approach to infer the driver behavior. The naturalistic driving data are automatically segmented in the first step and then assigned to the high-level discrete driving behavior. Taniguchi et al.\cite{taniguchi2016sequence} proposed to model the duration of segments of driving data by using a hierarchical Dirichlet process hidden semi-Markov model. A sequence of driving words is extracted from the data first, and then the driving letters which represent one typical discrete driving behavior are inferred. Inspired by the work of Taniguchi, Hamada et al.\cite{hamada2016modeling} extended the work to predict the future continuous operation behaviors using the estimated future state sequence. The states are segmented from the driving data by applying the beta process autoregressive hidden Markov model. Although the unsupervised methods have been successfully applied to segment the driving data based on their features, their focus is on operation behavior data, such as steering angle, brake pressure and accelerator pedal position, they do not concentrate on the particularity of trajectory segmentation. As for the vehicle trajectory analysis, Yao et al.\cite{yao2017road} established a lane change trajectory database to analysis the lane change behavior. The trajectories are extracted from the driving data by using the Support Vector Machines based classifier. As a result, the method is limited to extract the specific trajectory based on the labeled data.

\par In this paper, we apply heuristics to segment the time series data which is a general way for initial segmentation. Considering the particularity of trajectory segmentation, the zero crossing detection of the observed course deviation is served as the driving-specific heuristic because we prefer to maintain the integrity of the steering process. We apply probabilistic inference to find the correct segments and generate a weighted set of temporary segmented trajectories. The MPs which are learned from the temporary segmented trajectories are then utilized to improve the segmentation result by down-weighting the existed segments which are unlikely to belong to the current motion primitive library. When the iterative Expectation Maximization (EM) algorithm converges, we can get the suitable motion primitive library based on the optimal segmentation points. 

\par The main contributions of this paper are shown as follows.

\begin{itemize}		

\item Providing one possible method to represent the MPs which is both suitable for the vehicle motion planning task and imitation learning algorithm.

\item Implementing a probabilistic segmentation algorithm to solve the segmentation of unlabeled trajectory data and generation of point-point MPs in conjunction.

\end{itemize}	

\par The remains of this paper are organized as follows. Section \uppercase\expandafter{\romannumeral3} presents the DMPs and probabilistic segmentation methods that we use to solve this problem. Section \uppercase\expandafter{\romannumeral4} shows the experiment results for the representation and segmentation of MPs. Finally, the conclusions are given in Section \uppercase\expandafter{\romannumeral5}.

\section{Problem Statement}

According to a given set of observed vehicle motion trajectories \begin{math}  \boldsymbol{O} = \left\{ {{\boldsymbol{o}_1},{\boldsymbol{o}_2},...,{\boldsymbol{o}_n}} \right\} \end{math}, the goal of our work is to learn a set of underlying MPs \begin{math} \boldsymbol{M} = \left\{ {{\boldsymbol{m}_1},{\boldsymbol{m}_2},...,{\boldsymbol{m}_m}} \right\} \end{math} and generate the motion primitive library which can explain $\boldsymbol{O}$. 

\par The BIT intelligent vehicle (see in Fig.\ref{fig１(a)}) was selected as the data-collection platform. The driving situations include intersections, highways and ring roads in Beijing. The operations of the driver contain overtaking, lane keeping, lane changing and other typical driving operations.

\par The following variables are defined to represent the observed trajectory points and the generated motion primitive points.

\begin{itemize}

\item \begin{math} {\boldsymbol{o}_i}(t) = {[\Delta {\theta (t)},{v(t)}]^{\rm{T}}} \in {\mathbb{R}^{2 \times 1}}\end{math} is the definition of one observed vehicle trajectory point at time $t$, where $\Delta {\theta (t)}$ is the course deviation between time $t$ and time $t-1$, and $v(t)$ is the velocity of the trajectory point at time $t$.

\item \begin{math} {\boldsymbol{m}_i}(t) = [{v_{init}},\Delta {\theta _m}(t),\Delta {v_m}(t)] \in {\mathbb{R}^{3 \times 1}}\end{math} is the definition of the one motion primitive point at time $t$, where ${v_{init}}$ is the initial velocity of the selected motion primitive, \begin{math}	 \Delta {\theta _m}(t) = {\theta _t} - {\theta _{init}}	\end{math} and \begin{math} \Delta {v_m}(t) = {v_t} - {v_{init}} \end{math} are the generated course deviation and velocity deviation between the initial time and time $t$ separately. As for the steering, we only care about the relative change, so the initial course angle ${\theta _{init}}$ is not included in the definition.

\end{itemize}

\par Although the observed trajectories contain the longitudinal and lateral variables, only the lateral part $\Delta {\boldsymbol{\theta}}$ is served as the one-dimensional trajectory to determine the segmented points. For the selected trajectory that illustrated in Fig.\ref{fig１}, a group of possible cutting points which utilized to indicate the stars and ends of one segmented motion primitive is defined as $\boldsymbol{C}$.  We assume that the trajectory is over-segmented by the initial cutting points, i.e., the group of possible cutting points $\boldsymbol{C}$  includes both the true positive segments and false positive segments. As a result, each possible subset $\boldsymbol{d} \subseteq \boldsymbol{C}$ has to be regarded as a potential segmentation. Furthermore, the set of segmented trajectories that corresponds to $\boldsymbol{d}$ is defined as $\boldsymbol{S}$, where each segmented trajectory ${s_{i,j}} \in \boldsymbol{S}$ is defined by the starts and ends cutting points ${c_i},{c_j}$ (see in Fig.\ref{fig１(b)}).

\par The proposed method in this paper is trying to solve the following two challenges simultaneously: Determining the suitable segmented points ${\boldsymbol{d}^*} \subseteq \boldsymbol{C}$ from the possible cutting points and learning the MPs from the segmented trajectories to generate the library (see in Fig.\ref{fig１(c)}).

\section{Learning to Segment and Represent Motion Primitives}

\subsection{Initializing the Possible Cutting Points $\boldsymbol{C}$} 

As it is unfeasible to consider every time step of the observed trajectories’ points as the possible cutting points, we should restrict the number of the initial cutting points to an acceptable range. Therefore, some driving-specific heuristics should be used to determine the $\boldsymbol{C}$. Besides, as our method only considers the segmented points included in the $\boldsymbol{C}$ and removes the false positive segmented points, the possible cutting points should be a little bit over-segmented. Based on the above considerations, the zero crossing method is applied to initialize the segmented points in this paper, as explained in Fig.{\ref{fig１(b)}}. Only the lateral variables are chosen to calculate the possible segmented points because we want to guarantee that each steering process is always complete. As a result, the driver's steering behavior and the speed-fitting behavior in the steering process could be completely preserved.

\subsection{Representation of Motion Primitive}

There are three parameters which are applied to define the MPs. However, the initial velocity $v_{init}$ is always constant within the time duration of each motion primitive, so only the $\Delta {\boldsymbol{\theta _m}}$ and $\Delta {\boldsymbol{v_m}}$ need to be represented by the DMPs. 

As for the point to point movement, the canonical system is defined as
\begin{gather}
{\dot z_m} =  - \tau {\alpha _z}{z_m} \label{equ1}
\end{gather}
\par In order to enable the adaptation for the duration of MPs, the time scaling parameter $\tau  = 1/T$ is defined, where $T$ is the duration of the MPs. The ${z_m}$ is initially set to one and the parameter ${\alpha _z}$ is predefined to ensure that the ${z_m}$ converges to zero at time $T$ which is the end time of the chosen motion primitive. 

\par For driving movements, the learned DMPs are defined in the following form
\begin{gather}
{\boldsymbol{\dot v} = \tau {\alpha _y}({\beta _y}(\boldsymbol{g} - \boldsymbol{y}) - \boldsymbol{v}) + \tau \eta f({z_m})} \label{equ2} \\
{\boldsymbol{\dot y} = \tau \boldsymbol{v}} \label{equ3}
\end{gather}
\noindent where ${\boldsymbol{y} = [\Delta {\boldsymbol{\theta _m}},\Delta {\boldsymbol{v_m}}]}$, ${\boldsymbol{v} = [\Delta {\boldsymbol{\dot \theta _m}},\Delta {\boldsymbol{\dot v_m}}]}$, and the desired final position of the selected motion primitive is ${\boldsymbol{g} = [\Delta {\theta _m}(T),\Delta {v_m}(T)]}$. ${\mathop \eta  = (\boldsymbol{g} - \boldsymbol{y_0})}$ represents the amplitude, where ${\boldsymbol{y_0} = [\Delta {\theta _m}(1),\Delta {v_m}(1)]}$ is the initial position. The pre-defined constants ${\mathop {\alpha _y}}$ and ${\mathop {\beta _y}}$ are set to ensure the spring-damper system is critically damped.

\par In order to represent the nonlinear patterns in the randomly selected MPs, the transformation function is defined as 
\begin{gather}
f({z_m}) = \frac{{\sum\nolimits_{n = 1}^N {{\omega _n}{\psi _n}({z_m})} {z_m}}}{{\sum\nolimits_{n = 1}^N {{\psi _n}({z_m})}}}\label{equ4}
\end{gather}
\noindent where ${\psi _n}({z_m}) = exp( - {p_n}{({z_m} - {\mu _n})^2})$ is the Gaussian basis functions with ${\mu _n}$ represents the center and ${p_n}$ represents the bandwidth, ${\omega _n}$ is the adjustable parameters for the MPs and $N$ is the total number of adjustable parameters. The transformation function $f({z_m})$ changes the output results of the typical spring-damper model and therefore ensures the generation of arbitrarily shaped MPs. The influence of the transformation function $f({z_m})$ will vanish at the end of the MPs, because the $z$ converges to zero. This means the generated final position of the movement will converge to the desired final position without any influence of the non-linear function. That is the foundation for the representation of point-point MPs. 

\par As the predefined parameters ${\alpha _z}$, ${\alpha _y}$ and ${\beta _y}$ are always fixed in the representation of motion primitive, the representation parameters for the lateral MPs and longitudinal MPs ${\boldsymbol{\omega _n}} = [{\omega _\theta }(n),{\omega _v}(n)]$ are needed to be learned from the observed trajectories.

\par The problem of the learning method is to infer the adjustable parameters ${\boldsymbol{\omega _\theta} },{\boldsymbol{\omega _v}}$ in order to minimize the errors between the represented and the demonstrated trajectory. Therefore, the problem can be solved directly by a regression algorithm\cite{ijspeert2002movement}. As the solution process is almost the same between the longitudinal and lateral part, only the detailed method for learning lateral MPs from the demonstrated trajectory is shown in Algorithm 1.

\begin{algorithm}  
\caption{Learning of lateral motion primitive}  
\label{alg:A}  
\begin{algorithmic}  
\REQUIRE ${M_t} = [\Delta {\theta _m(t)},\Delta {\dot \theta _m(t)},\Delta {\ddot \theta _m(t)}]$, $t \in \{ 1,...,T\}$
\FOR {each parameter ${\omega _\theta }(n)$}
\STATE Set predefined parameters ${\alpha _z}$,${\alpha _y}$,${\beta _y}$,$\tau  = 1/T$,$g = \Delta {\theta _T}$
\STATE Calculate ${z_m}(t)$ by integrating ${\dot z_m} =  - \tau {\alpha _z}{z_m}$ for all time duration
\STATE Calculate the deviation value ${f^{dev}}(t)$ by using the input data
\STATE ${f^{dev}}(t) = \Delta {\ddot \theta _m}(t)/{\tau ^2} - {\alpha _y}({\beta _y}(g - \Delta {\theta _m}(t)) - \Delta {\dot \theta _m}(t))$
\STATE Calculate the Gaussian basis functions
\STATE $\psi _t^n({z_m}(t)) = exp( - {p_n}{({z_m}(t) - {\mu _n})^2})$
\STATE Set matrices $\boldsymbol{z} = {\left[ {{z_m}(1),...,{z_m}(T)} \right]^{\rm{T}}}$, $\boldsymbol{\psi}  = diag(\psi _1^n,...,\psi _T^n)$ and $\boldsymbol{f^{dev}} = {[{f^{dev}}(1),...,{f^{dev}}(T)]^{\rm{T}}}$
\STATE Compute ${\omega _\theta }(n)$ by locally weighted regression
\STATE ${\omega _\theta }(n) = {({\boldsymbol{z}^{\rm{T}}}\boldsymbol{\psi} \boldsymbol{z})^{ - 1}}{\boldsymbol{z}^{\rm{T}}}\boldsymbol{\psi} \boldsymbol{f^{dev}}$
\ENDFOR
\end{algorithmic}  
\end{algorithm}  

\par By using the Algorithm 1 we can calculate the ${\boldsymbol{\omega} _\theta },{\boldsymbol{\omega} _v}$ for every segmented MPs,then the representation parameter set and adjustment parameter set for each MPs are ${\boldsymbol{\phi} _m} = [{v_{init}},{\boldsymbol{\omega} _\theta },{\boldsymbol{\omega} _v}]$ and ${\boldsymbol{\gamma} _m} = [\boldsymbol{g},T]$ separately.

\begin{figure*}
\centering
\subfloat[Sharp turn]{
\begin{minipage}[t]{0.95\textwidth}
\centering
\includegraphics[scale=0.4]{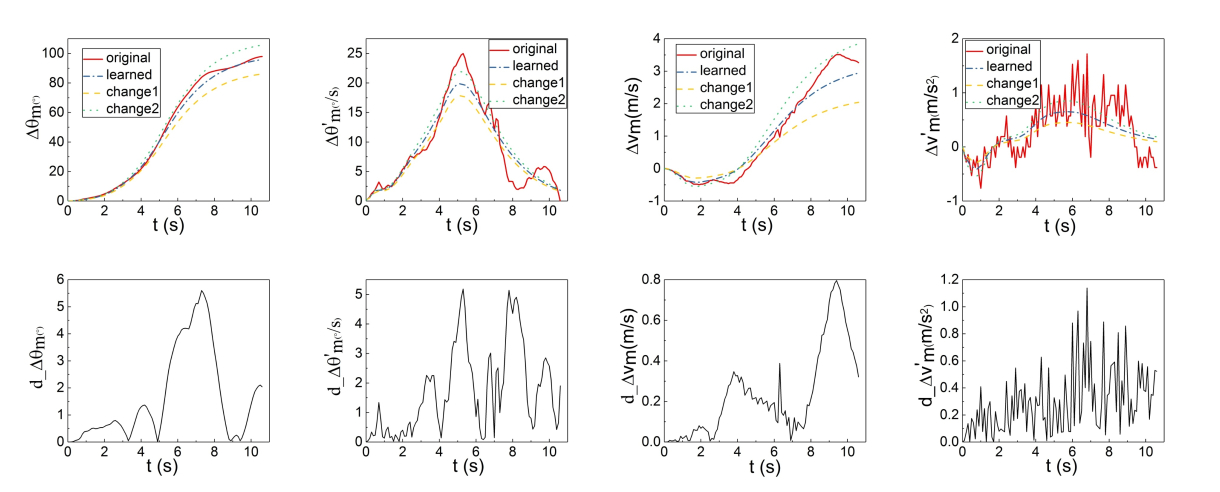}
\end{minipage}
\label{fig2(a)}
}

\subfloat[Lane changing]{
\begin{minipage}[t]{0.95\textwidth}
\centering
\includegraphics[scale=0.4]{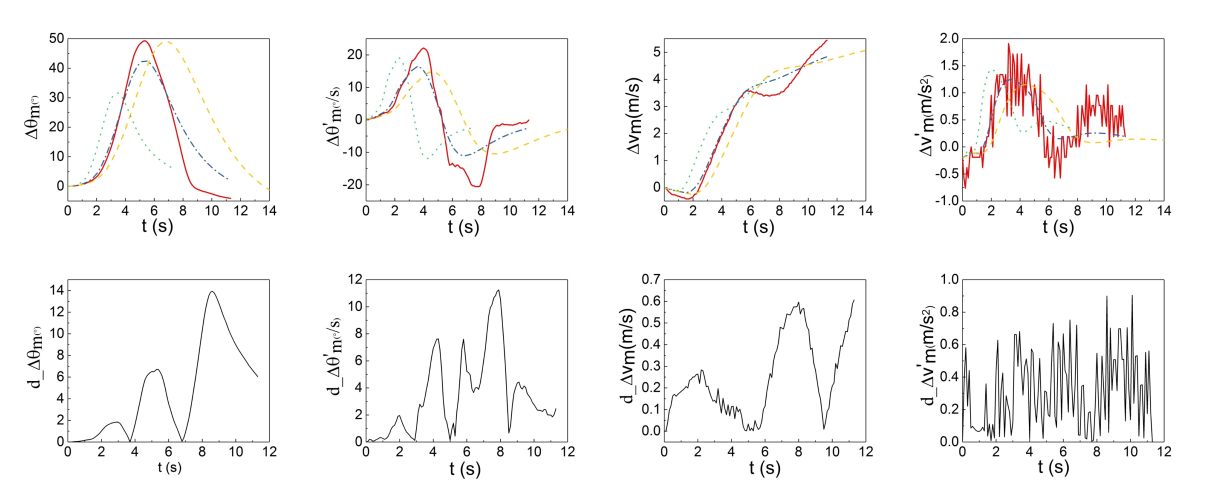}
\end{minipage}
\label{fig2(b)}
}
\caption{This figure illustrates the accuracy of our MPs representation method and the adaptation of the learned MPs to the change of final goal position and time duration which are essential for both the imitation learning and motion planning. Two typical driving operations were selected to evaluate the performance of representation method. In each situation, the $\Delta {\boldsymbol{\theta} _m},\Delta {\boldsymbol{\dot \theta _m}},\Delta {\boldsymbol{v_m}},\Delta {\boldsymbol{\dot v_m}}$ and the corresponding deviations were presented in order. Situation a concentrated on the goal value change of $\Delta {\theta _m}(T),\Delta {v_m}(T)$ and situation b demonstrated the change in time duration $T$. The learned parameter set ${\boldsymbol{\phi} _m}$ remains unchanged in the adjustment. It could be found that the overall shape of the trajectory is not changed during the adjustment.}
\label{fig2}
\end{figure*}

\subsection{Probabilistic Segmentation Method}

Each motion primitive $\boldsymbol{m} \in \boldsymbol{M}$ is represented by the parameter set $\boldsymbol{\phi} _m$. Therefore, the library of MPs is defined as a mixture of every single motion primitive, in order to calculate the probability of a selected segmented trajectory given the entire library.
\begin{gather}
p(\boldsymbol{s}|\boldsymbol{\Phi} ) = \sum\nolimits_{m = 1}^M {{\lambda _m}p(\boldsymbol{s}|{\boldsymbol{\phi} _m})} \label{equ5} \\
\boldsymbol{\Phi}  = \{ ({\lambda _1},{\boldsymbol{\phi}_1}),...,({\lambda _M},{\boldsymbol{\phi} _M})\} \label{equ6}
\end{gather}
\noindent where ${\lambda _m}$ is the mixture coefficient for each motion primitive in the library.

\par We assume that each observed trajectory $\boldsymbol{o} \in \boldsymbol{O}$ is represented by a parameterized model.
\begin{gather}
\boldsymbol{o} \sim p(\boldsymbol{o}|\boldsymbol{\Phi} ,{\boldsymbol{S}^*}) = \prod\nolimits_{\boldsymbol{s} \in {\boldsymbol{S}^*}} {p(\boldsymbol{s}|\boldsymbol{\Phi} )} \label{equ7}
\end{gather}
\par The latent variable ${\boldsymbol{S}^*}$ can be integrated out by using Eq.\ref{equ8}.
\begin{gather}
p(\boldsymbol{o}|\boldsymbol{\Phi} ) = \sum\nolimits_{\boldsymbol{S} \in \boldsymbol{D}} {p(\boldsymbol{S})} \prod\nolimits_{\boldsymbol{s} \in \boldsymbol{S}} {p(\boldsymbol{s}|\boldsymbol{\Phi} )}\label{equ8}
\end{gather}
\par The most likely parameter model $\boldsymbol{\Phi}$ of motion primitive library is computed by maximizing the likelihood.
\begin{gather}
{\boldsymbol{\Phi ^*}} = \arg \mathop {\max }\limits_\Phi  \sum\nolimits_{\boldsymbol{o} \in \boldsymbol{O}} {\log p(\boldsymbol{o}|\boldsymbol{\Phi} )} \label{equ9}
\end{gather}
\par We apply the EM algorithm to optimize this likelihood and the auxiliary function is defined as 
\begin{gather}
\boldsymbol{\Phi}  = \arg \mathop {\max }\limits_\Phi  Q(\boldsymbol{\Phi} ,\boldsymbol{\Phi} ') \label{equ10} \\
Q(\boldsymbol{\Phi} ,\boldsymbol{\Phi} ') = \sum\nolimits_{\boldsymbol{o} \in \boldsymbol{O}} {\sum\nolimits_{\boldsymbol{S} \in \boldsymbol{D}} {p(\boldsymbol{S}|\boldsymbol{o},\boldsymbol{\Phi} ')\log ({p_s}p(\boldsymbol{o}|\boldsymbol{\Phi} \boldsymbol{S})} )} \label{equ11}
\end{gather}
\noindent where ${\boldsymbol{\Phi} '}$ is defined as model parameter set which is calculated from the previous iteration. The ${\alpha _s} = \sum\nolimits_{\boldsymbol{S} \in \boldsymbol{D}} {p(\boldsymbol{S}|\boldsymbol{o},\boldsymbol{\Phi} ')}$ indicates the probability of each segment to be the correct segmentation. The prior ${p_s}$ is defined as a product of priors over segments.
\begin{gather}
{p_s} = {p_c}\prod\limits_{s \in S} {p({c_s})} \label{equ12} \\
p({c_s}) = {(1 - {p_c})^{{c_s}}}{p_c} \label{equ13}
\end{gather}
\noindent where ${c_s}$ is the total number of cutting points that the possible segmented trajectory $s$ spans over. The parameter $0 < {p_c} < 1$ indicates how suitable it is that the cutting points which determine the segmented trajectory s are the true positive. When ${p_c} < 0.5$ indicates that longer segmented trajectory is preferred. On the contrary, shorter segments are preferable. 

\par By using the function defined in Eq.\ref{equ11}, the weighting parameter $\alpha _s$ is computed from the E-step and the Gaussian Mixture Models (GMMs) based parameter set ${\boldsymbol{\Phi} '}$  is estimated in the M-Step. Once the algorithm converges, we can simultaneously get the correct segmented points and the learned GMMs to represent the parameter set $\boldsymbol{\Phi}$ of motion primitive library.

\subsection{EM-GMM Segmentation Method}
	The EM-GMM segmentation method is selected as the baseline method to compare with the probabilistic segmentation method. The initial possible cutting points and the DMP are not applied in this method.
	
\par In the training process of the EM-GMM segmentation method, the feature dataset ${\boldsymbol{x}_{p}}(t)$ is chosen as 
\begin{gather}
{\boldsymbol{x}_{p}}(t) = {\boldsymbol{o}_i}(t) = [\Delta {\theta (t)},{v(t)}] \label{equ14}
\end{gather}

\par The GMM is represented as
\begin{gather}
G(\boldsymbol{x};\boldsymbol{\beta}) = \sum\limits_{i = 1}^K {{p_i}{g_i}(\boldsymbol{x};{\boldsymbol{\mu} _i},{\boldsymbol{\Sigma} _i})}  \label{equ15}\\
{g_i}(\boldsymbol{x};{\boldsymbol{\mu} _i},{\boldsymbol{\Sigma} _i}) = \frac{1}{{\sqrt {{{(2\pi )}^d}\left| {{\boldsymbol{\Sigma} _i}} \right|} }}{e^{ - \frac{1}{2}{{(x - {\boldsymbol{\mu} _i})}^T}{\boldsymbol{\Sigma} _i}^{ - 1}(x - {\boldsymbol{\mu} _i})}} \label{equ16}
\end{gather}

\noindent where $\boldsymbol{x}$ is a set of d-dimension sequence, $\boldsymbol{x} = \left\{ {{\boldsymbol{x}_i}} \right\}_{i = 1}^n$ with ${x_i} \in {\mathbb{R}^{d \times 1}}$, ${\mu _i} \in {\mathbb{R}^{d \times 1}}$ and ${\Sigma _i} \in {\mathbb{R}^{d \times d}}$ are mean vector and covariance vector of a single Gaussian Model, $p_i$ is the weight coefficient of a single Gaussian Model and $\sum\nolimits_{i = 1}^n {{p_i} = 1}$, $\boldsymbol{\beta}  = \left\{ {{\boldsymbol{\beta} _i}} \right\}_{i = 1}^n$ with ${\boldsymbol{\beta} _i} = \left\{ {{\boldsymbol{\mu} _i},{\boldsymbol{\Sigma} _i},{p_i}} \right\}$. Maximum Likelihood Estimation of the model parameters is achieved iteratively using the Expectation-Maximization (EM) algorithm.

\par After the model has been trained by applying the chosen path features, the path point sequences can be labeled by the trained model.Therefore, the trajectories are segmented by the learned labels, and the path primitives are based on the ${\boldsymbol{x}_{p}}(t)$ not the parameter set $\boldsymbol{\Phi}$ which is used in the probabilistic segmentation method.

\section{Experiments and Results}

\subsection{Data Collection System}

\par The proposed method in this paper mainly focuses on the driver’s driving skills and does not involve the comparison of the multi-driver driving styles. As a result, only one driver’s driving data was selected from the 81 hours driving data to train the model in this paper.

\par The trajectory dataset which was mainly used in this paper was collected by the equipped OxTs integrated navigation unit. Course angle, horizontal velocity and timestamp of the collection were provided by this sensor.

\subsection{Evaluation of Typical Motion Primitives}

\par There are a total of 26 basic MPs in the library. However, it was meaningless to demonstrate every motion primitive in the library. As a result, only two typical MPs were selected to evaluate the performance of our representation method. The two selected typical motion primitives were sharp turn and lane changing motion primitive. The reasons we chose these two motion primitives is that the two common steering processes contain typical longitudinal-lateral cooperation relationship which is the crucial factor that the learned motion primitives would like to represent.

\par In order to evaluate the accuracy of our representation method, the position and velocity deviations were chosen as the performance indexes. The position deviation ${d_{pos}}$ can be described as
\begin{gather}
{d_{pos}} = \left| {{y_d} - {y_l}} \right| \label{equ19}
\end{gather}

\par The velocity deviation ${d_{vel}}$ can be calculated by 
\begin{gather}
{d_{vel}} = \left| {{v_d} - {v_l}} \right| \label{equ20}
\end{gather}

\noindent where ${y_{d/l}}$ and ${v_{d/l}}$ are the motion primitive data of the position and velocity respectively, the subscript $d$ and $l$ represent the demonstrated and learned data respectively.

\par We also presented the adaptability to the change of goal position and time duration (see in Fig.\ref{fig2}).

\subsection{Comparison of the Trajectory Segmentation}

\begin{figure}
\centering
\subfloat[Low-speed with sharp turns]{
\begin{minipage}[t]{0.45\textwidth}
\centering
\includegraphics[scale=0.072]{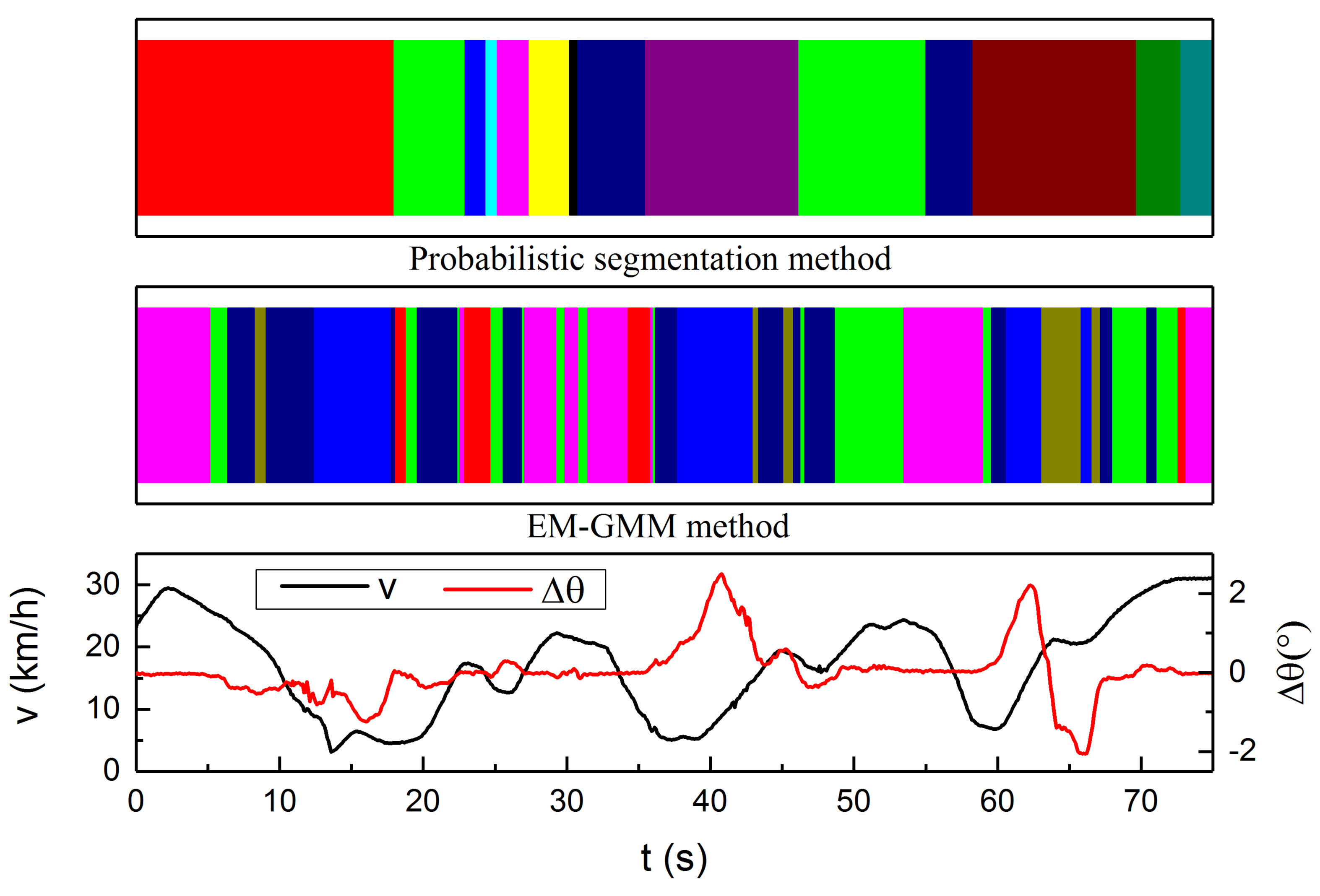} 
\end{minipage}
\label{fig3(a)}
}

\subfloat[High-speed with direction corrections]{
\begin{minipage}[t]{0.45\textwidth}
\centering
\includegraphics[scale=0.072]{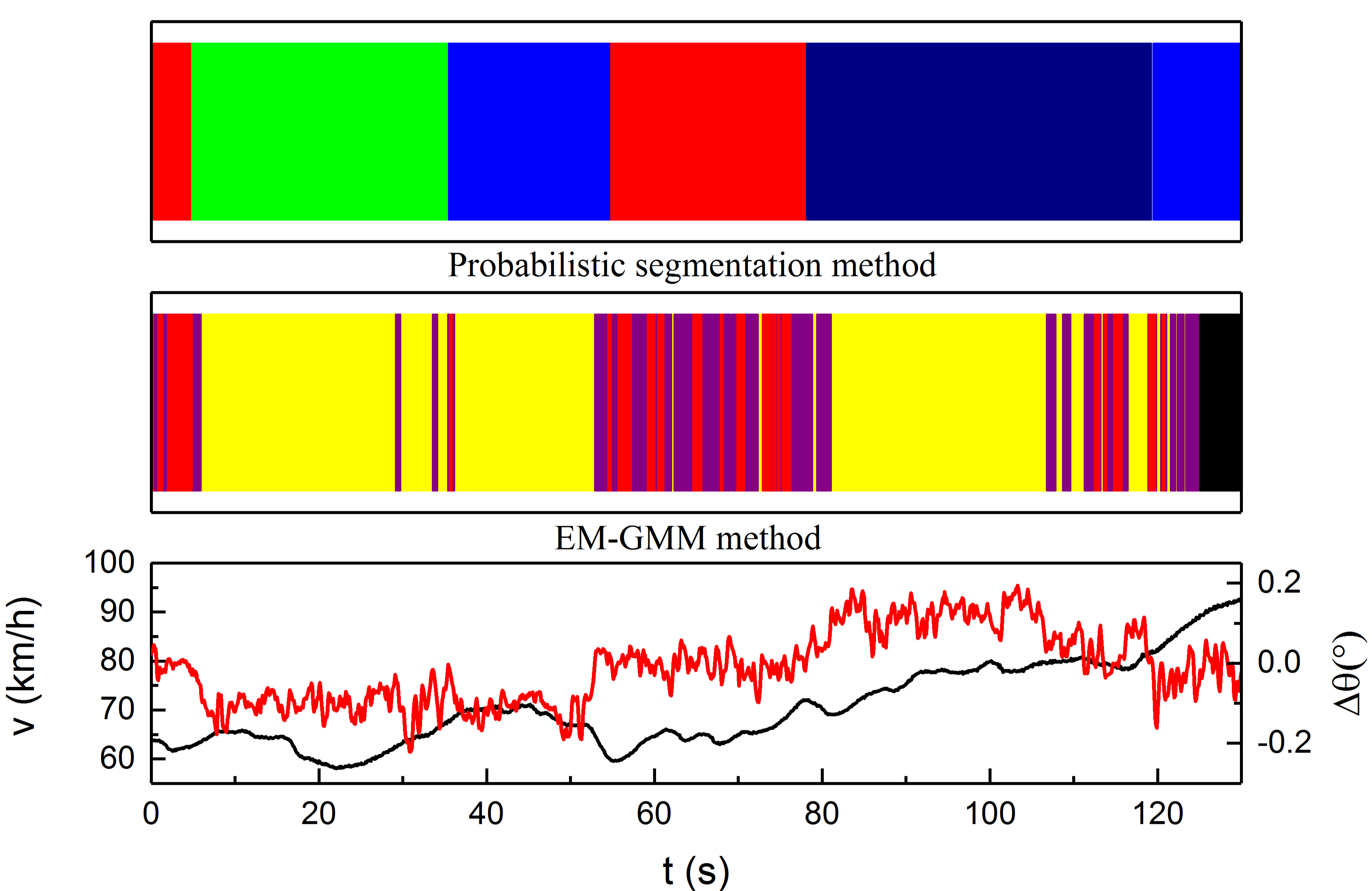} 
\end{minipage}
\label{fig3(b)}
}
\caption{The segmentation results of two typical driving situations. In each subgraph, the segmentation result of the probabilistic segmentation method, the segmentation result of the EM-GMM method and the observed path data were presented in order. The EM-GMM was chosen as the representative algorithm of unsupervised learning approaches which can automatically extract MPs from driving data without any initial heuristics. Different colors within each segmentation result illustrated different types of MPs, but there was no association between colors in different result graphs. Some of the segments of the EM-GMM method were too short to be seen in the plot.}
\label{fig3}
\end{figure}

In order to demonstrate the benefits of optimizing both the trajectory segmentation and the motion primitive library iteratively, we applied the EM-GMM method as a baseline. The baseline method clusters the dataset $\boldsymbol{O}$ directly in order to get the identified trajectory type and segment the trajectory based on the identified type. The segmentation comparison results based on these two methods were shown in Fig.\ref{fig3}.

\par The performance of segmentation method was evaluated by the number of active segmented points $n_2$ (see in Table.\ref{table_1}), and less segmentation points mean more compact representation of the observed trajectory. 
\begin{table}[h]
 \centering
 \begin{tabular}{c c c c}
  \toprule
  Type & $n_1$ & $n_2$(Probs) & $n_2$(EM-GMM) \\ 
  \midrule
  1 & 37 & 14 & 42 \\
  2 & 82 & 8 & 94\\
  \bottomrule
 \end{tabular}
 \caption{The segmentation results of probabilistic segmentation (Probs) results and EM-GMM method. The type 1 indicates the driving situation of low-speed with sharp turns, and type 2 indicates the driving situation of high-speed with direction corrections. $n_1$ is the initial cutting points of the selected trajectory, and $n_2$ is the number of active segmented points after the learning process.}
 \label{table_1}
\end{table}

\subsection{Discussion}

The benefit of our approach becomes quite obvious when comparing the probabilistic segmentation approach with the reference EM-GMM method, shown in Fig.\ref{fig3}. Despite achieving comparable segmentation results, the proposed method can achieve more meaningful segment results which correspond to the realistic driving situations compared with the baseline method. The reason is that the probabilistic segmentation is based on the initial cutting points which are determined by the prior knowledge of the motion primitive settings. The purpose of the proposed method is to merge the positive false cutting points instead of directly finding the positive true segment points. However, the EM-GMM method or other similar learning methods which automatically extract MPs from driving data without any prior knowledge are suitable for many applications, but they may not be the proper methods to segment the trajectory for the motion planning applications. As for the MPs which are used for the motion planning tasks, some heuristics can and must be made previously for meaningful segmentation and reusing of segmented MPs.

\par Besides, as our method takes advantage of the mutual dependencies between the segmentation and representation, so the representation method is another crucial factor. As shown in Fig.\ref{fig2}, the proposed representation method can be adapted to the change of final goal position and time duration with the same representation parameter set ${\boldsymbol{\phi} _m}$. It is essential for both the representation and reusing of the learned motion primitive. As for the representation, the trajectory can be represented by the same set of parameters as long as the basic shape of the trajectory is the same. This will greatly reduce the number of possible primitive types and enhance the effectiveness of the segmentation algorithm. As for reusing of MPs to motion planning application, we can achieve different target points with the same overall shape of the trajectory by only changing the adjustment parameter set ${\boldsymbol{\gamma} _m}$. The representation parameter set ${\boldsymbol{\phi} _m}$ which is served as the identification of different shape types remains unchanged during the adaptation. This property is the basis of how DMPs can be used as a representation method for the MPs library. Furthermore, the representation method is also robustness to sudden perturbations and can resist fluctuations near zero. However, it is unable to represent emergency driving behaviors which contain sudden changes in the values of the trajectory. As a result, all the emergency driving situations are removed from the training dataset manually with the help of collected camera data.

\section{Conclusion}

In this paper, we proposed a new framework which can segment the unlabeled trajectories and learn a motion primitive library in conjunction. The established motion primitive library is used to infer the true positive segment points from the possible initial cutting points which are chosen based on the zero detection of the observed course deviation. Then the inferred segmentations, in turn, help to optimize the MPs in the library by utilizing the mutual dependency between the segmentation and representation. We evaluated the representation accuracy and adaptation of the learned MPs by using the driving data collected from the BIT intelligent vehicle platform, and we also compare our segmentation method with one of the typical unsupervised learning approaches. We found that our approach is able to learn more compact and meaningful MPs compared with the unsupervised learning method without any driving-specific heuristics, and the learned MPs which can adapt to the change of goal position and time duration are suitable for both the imitation learning and motion planning. The proposed method can be applied to learn the human driving skills from the collected driving data in a more efficient way, and the learned MPs library can be combined with the traditional motion planning method to generate personalized motion pattern.

\par In future work, we aim at adjusting the representation method of the MPs to make it more suitable for driving tasks, combining the proposed approach with some stable motion planning methods and optimizing the MPs online.

\bibliographystyle{unsrt}
\bibliography{root}

\end{document}